# Versatile Land Navigation Using Inertial Sensors and Odometry: Self-calibration, In-motion Alignment and Positioning

Yuanxin Wu

**Abstract—Inertial measurement unit (IMU) and odometer have been commonly-used sensors for autonomous land navigation in the global positioning system (GPS)-denied scenarios. This paper systematically proposes a versatile strategy for self-contained land vehicle navigation using the IMU and an odometer. Specifically, the paper proposes a self-calibration and refinement method for IMU/odometer integration that is able to overcome significant variation of the misalignment parameters, which are induced by many inevitable and adverse factors such as load changing, refueling and ambient temperature. An odometer-aided IMU in-motion alignment algorithm is also devised that enables the first-responsive functionality even when the vehicle is running freely. The versatile strategy is successfully demonstrated and verified via long-distance real tests.**

## I. Introduction

Determination of attitude, velocity and position is an essential task in many land applications, such as mobile mapping and navigation, pipeline pigging, and borehole surveying. Dead-reckoning inertial navigation by triads of gyroscopes and accelerometers is an old but wildly employed means among many [2]. Inertial navigation system (INS) typically has high-bandwidth output but short-duration stability because integration of noise-contaminated inertial measurements inevitably leads to a unbounded drift. In practice, other relative or absolute sensors with complementary properties are introduced to mitigate the error drift, such as the global positioning system (GPS), cameras, odometers and Doppler radars. Specifically, GPS has received overwhelming attention as an absolute updating sensor in recent decades because of its popularity and low cost. GPS measurement accuracy is time-irrelevant, yet prone to signal blockage, jamming and spoofing in urban and hostile environment. Camera is also a promising choice despite its tight dependence on easily-identified features with known positions on the path to go.

For land navigation of interest hereby, an odometer is a cost-effective and conveniently-deployed sensor, widely equipped in modern land vehicles for the antilock braking system. As a relative positioning sensor, it measures speed or incremental distance along the vehicle trajectory. Reliability of the odometer outputs depends on land surface conditions and vehicle maneuvers, deteriorating if relative slippage occurs between the tires and contact surface when braking sharply or driving on soft land. On the other hand, an odometer is a very reliable sensor on benign land surfaces and has accurate short-distance stability that is independent of running time. As compared to airborne applications, driving on land is a constrained motion in that the vehicle moves forward tangential to its trajectory and the velocity in the plane perpendicular to the moving direction is zero. It is well known as the nonholonomic constraint, which was first introduced into land navigation by an Australian robotics group [3] and has received many interests in recent years [3-8]. It was shown that the nonholonomic constraint, alone or along with an odometer, contributes much to the limitation of error drift of pure inertial navigation and turns out to be very useful prior model information for land navigation.

To fully use the priori nonholonomic constraint and the odometer information, we first of all should know the spatial misalignment of the inertial measurement unit (IMU) relative to the vehicle. For real systems, however, the vehicle frame misaligns the IMU frame in attitude and position, which has seldom been seriously considered in the literature or by the well-known navigation product providers, e.g., iMAR and NovAtel. The IMU-vehicle attitude and/or position misalignment is a common problem in any real system involving multiple sensors, which, if not considered properly, can noticeably decay the system accuracy [9]. Usually, the position misalignment (called the lever arm) is compensated with quantities that are manually measured on the spot [6, 7], but it is not easy and even very difficult to measure the spatial misalignment, i.e., the misaligning angles, with enough accuracy to reliably apply the prior nonholonomic constraint and/or the odometer outputs. Additionally, the misalignment may have non-negligible variations due to such inevitable factors as load changing, refueling and ambient temperature. In our previous work [10], the feasibility of self-calibration was first analyzed from the perspective of nonlinear global observability, moderate sufficient conditions were obtained

This work was supported in part by the Fok Ying Tung Foundation (131061), National Natural Science Foundation of China (61174002, 61422311), the Foundation for the Author of National Excellent Doctoral Dissertation of People's Republic of China (FANEDD 200897) and Program for New Century Excellent Talents in University (NCET-10-0900).

Authors' address: Yuanxin Wu (corresponding author), School of Aeronautics and Astronautics, Central South University, Changsha, Hunan, China, 410083. Tel: 086-0731-88877132, E-mail: (yuanx_wu@hotmail.com).

and guidelines for selecting a practical calibration path were proposed. Our subsequent work [11] proposed an estimation scheme to automatically calibrate the IMU-vehicle attitude/position misalignment using outputs from the IMU and an uncalibrated odometer, as well as the nonholonomic constraint and GPS aiding. Since GPS information was incorporated therein, the GPS antenna lever arm had to be considered as we discussed in [12]. Given a good calibration of IMU-vehicle misalignment, the nonholonomic constraint along with an odometer forms a three-dimensional vehicle speedometer, a valuable aid to INS attitude initialization while on the run. In-motion INS alignment will be an advantageous functionality for special application scenarios.

An ultimate solution to autonomous land navigation using only inertial sensors and odometry is envisioned in this paper, reducing absolute position fixes to the minimum requirement. The solution comprises self-calibration of the parameters of the two sensor suites (IMU and odometer/nonholonomic constraint), in-motion INS alignment, and accurate positioning without absolute fixes. The main contributions of the paper are two-fold. Firstly, a self-calibration and refinement method is proposed for INS/odometer land navigation, which is able to not only estimate the IMU-vehicle misalignment from scratch but also mitigate significant parameter variation. Secondly, an odometer-aided in-motion alignment algorithm is raised that serves the autonomous strategy of self-contained land navigation. The proposed algorithms are successfully demonstrated in 200 km tests. Hopefully, the paper will benefit applications that use GPS during the signal outages and those using a camera when known features cannot be identified. The contents are organized as follows. Section II mathematically describes the problem of INS/odometer land navigation. In view of difficulty of the problem as a whole, Section III divides it into two separate but tightly coupled sub-problems and analyzes their observability properties, and Section IV presents the algorithms and implementation to solve the two sub-problems. Test results are reported in Section V and conclusions are drawn in Section VI.

## II. Problem Description

Let us consider a land vehicle equipped with an odometer and a full IMU. Figure 1 illustrates a front-wheel steering vehicle, the IMU, and the vehicle frame $A$ that originates at the center of the rear axle, point o, with $x$-axis directing forward, $y$-axis right along the rear axle and $z$-axis downward.

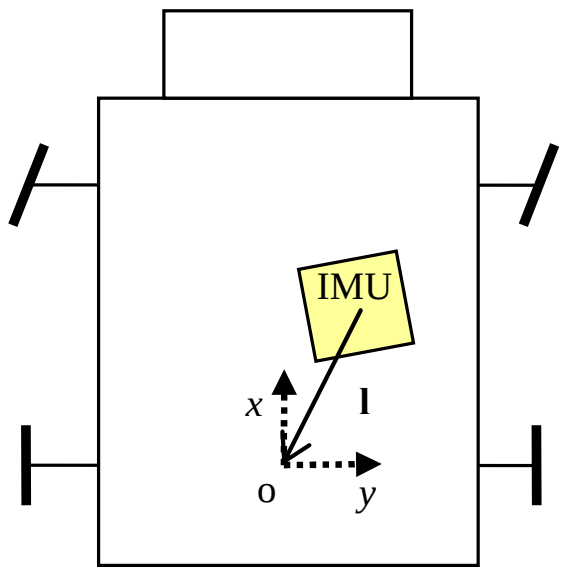

Figure 1. IMU and the vehicle frame (top view). The $z$ axes points downward.

Note that the vehicle frame has three other alternative definitions, e.g., a new vehicle frame can be obtained by reversing $y$-axis and $z$-axis to their opposites [10]. The IMU is rigidly fixed to the vehicle, with the body frame $B$ implicitly defined by the configuration of three gyroscopes/accelerometers. Practically, the body frame misaligns the vehicle frame in attitude and translation. Denote by $\mathbf{l}$ the translational displacement of the vehicle frame relative to the body frame, alternatively called as the odometer lever arm relative to the IMU.

Denote by $N$ the local level reference frame, by $I$ the inertially non-rotating frame, and by $E$ the Earth frame. The navigation (attitude, velocity and position) rate equations in the reference $N$-frame are respectively well known as [2, 13, 14]

$$\dot{\mathbf{C}}_b^n = \mathbf{C}_b^n \left(\boldsymbol{\omega}_{nb}^b \times\right), \quad \boldsymbol{\omega}_{nb}^b = \boldsymbol{\omega}_{ib}^b - \mathbf{b}_g - \mathbf{C}_n^b \boldsymbol{\omega}_{in}^n \tag{1}$$

$$\dot{\mathbf{v}}^n = \mathbf{C}_b^n \left(\mathbf{f}^b - \mathbf{b}_a\right) - \left(2\boldsymbol{\omega}_{ie}^n + \boldsymbol{\omega}_{en}^n\right) \times \mathbf{v}^n + \mathbf{g}^n \tag{2}$$

$$\dot{\mathbf{p}} = \mathbf{R}_c \mathbf{v}^n \tag{3}$$

$$\dot{\mathbf{b}}_g = 0 \tag{4}$$

$$\dot{\mathbf{b}}_a = 0 \tag{5}$$

where $\mathbf{C}_b^n$ denotes the attitude matrix from the body frame to the reference frame, $\mathbf{v}^n$ the velocity relative to the Earth, $\boldsymbol{\omega}_{ib}^b$ the error-contaminated body angular rate measured by gyroscopes in the body frame, $\mathbf{f}^b$ the error-contaminated specific force measured by accelerometers in the body frame, $\boldsymbol{\omega}_{ie}^n$ the Earth rotation rate with respect to the inertial frame, $\boldsymbol{\omega}_{en}^n$ the angular rate of the reference frame with respect to the Earth frame, $\boldsymbol{\omega}_{nb}^b$ the body angular rate with respect to the reference frame, and $\mathbf{g}^n$ the gravity vector. The $3\times 3$ skew symmetric matrix $(\cdot\times)$ is defined so that the cross product satisfies $\mathbf{a}\times\mathbf{b} = (\mathbf{a}\times)\mathbf{b}$ for arbitrary two vectors. The gyro bias $\mathbf{b}_g$ and the accelerometer bias $\mathbf{b}_a$ are taken into considerations approximately as random constants.

The position $\mathbf{p} \triangleq \begin{bmatrix}\lambda & L & h\end{bmatrix}^T$ is described by the angular orientation of the reference frame relative to the Earth frame, commonly expressed as longitude $\lambda$ and latitude $L$, and the height above the Earth surface $h$. $\mathbf{R}_c$ is the local curvature matrix that is a function of the current position. In the context of a specific local level frame choice, e.g., North-Up-East, $\mathbf{v}^n = \begin{bmatrix}v_N & v_U & v_E\end{bmatrix}^T$, the local curvature matrix is explicitly expressed as a function of current position

$$\mathbf{R}_c = \begin{bmatrix} 0 & 0 & \dfrac{1}{(R_E + h)\cos L} \\ \dfrac{1}{R_N + h} & 0 & 0 \\ 0 & 1 & 0 \end{bmatrix} \tag{6}$$

where $R_E$ and $R_N$ are respectively the transverse radius of curvature and the meridian radius of curvature of the reference ellipsoid. The specific expression of $\mathbf{R}_c$ will be different for other local level frame choices but it does not hinder from understanding the main idea of this paper.

In our configuration, the odometer measures the forward speed of point *o* in Fig. 1. The odometer outputs satisfy

$$y_{odo} = f \cdot \mathbf{e}_1^T \mathbf{C}_b^a \left( \mathbf{C}_n^b \mathbf{v}^n + \boldsymbol{\omega}_{eb}^b \times \mathbf{l}^b \right) \tag{7}$$

where *f* is a unknown coefficient factor of the odometer and $\mathbf{e}_i$ is a unit three-dimensional column vector with the $i^{\text{th}}$ element being 1 and others zero. The matrix $\mathbf{C}_b^a$ denotes the attitude misalignment from the body frame to the vehicle frame and $\mathbf{l}^b$ is the odometer lever arm expressed in the body frame. Suppose the rotation sequence from the body frame to the vehicle frame is first around *y*-axis (yaw angle, $\psi$ ), followed by *z*-axis (pitch angle, $\theta$) and by *x*-axis (roll angle, $\phi$), the attitude misalignment $\mathbf{C}_b^a$ is encoded by Euler angles as

$$\mathbf{C}_b^a = \begin{bmatrix} \cos\theta\cos\psi & \sin\theta & -\cos\theta\sin\psi \\ \sin\phi\sin\psi - \cos\phi\cos\psi\sin\theta & \cos\phi\cos\theta & \cos\psi\sin\phi + \cos\phi\sin\theta\sin\psi \\ \cos\phi\sin\psi + \cos\psi\sin\phi\sin\theta & -\cos\theta\sin\phi & \cos\phi\cos\psi - \sin\phi\sin\theta\sin\psi \end{bmatrix} \tag{8}$$

For land vehicular applications controlled by the nonholonomic constraint, the vehicle velocity in the plane perpendicular to the moving direction is zero

$$\mathbf{y}_{nhc} = \begin{bmatrix} 0 \\ 0 \end{bmatrix} = \begin{bmatrix} \mathbf{e}_2^T \\ \mathbf{e}_3^T \end{bmatrix} \mathbf{C}_b^a \left( \mathbf{C}_n^b \mathbf{v}^n + \boldsymbol{\omega}_{eb}^b \times \mathbf{l}^b \right) \tag{9}$$

which can be regarded as "virtual measurements". Merging (6) and (7) get the total measurement equation as

$$\mathbf{y} = \begin{bmatrix} y_{odo} \\ \mathbf{y}_{nhc} \end{bmatrix} = \operatorname{diag}\left( \begin{bmatrix} f & 1 & 1 \end{bmatrix} \right) \mathbf{C}_b^a \left( \mathbf{C}_n^b \mathbf{v}^n + \boldsymbol{\omega}_{eb}^b \times \mathbf{l}^b \right) \tag{10}$$

The roll angle along *x*-axis is totally irrelevant to the measurement [10], or alternatively unobservable, because using (8) the above equation can be rewritten as

$$\begin{aligned} \mathbf{C}_n^b \mathbf{v}^n + \boldsymbol{\omega}_{eb}^b \times \mathbf{l}^b &= \mathbf{C}_a^b \operatorname{diag}^{-1}\left( \begin{bmatrix} f & 1 & 1 \end{bmatrix} \right) \begin{bmatrix} y_{odo} \\ 0 \\ 0 \end{bmatrix} \\ &= \mathbf{C}_a^b \begin{bmatrix} y_{odo}/f \\ 0 \\ 0 \end{bmatrix} = \frac{y_{odo}}{f} \begin{bmatrix} \cos\theta\cos\psi \\ \sin\theta \\ -\cos\theta\sin\psi \end{bmatrix} \end{aligned} \tag{11}$$

From here on, we disregards the unobservable roll angle in the attitude misalignment, by assuming $\phi = 0$ in (8).

In addition to those navigation parameters in (1)-(5), the attitude/position misalignment and the odometer coefficient factor need also to be calibrated, which here are treated as unknown but random constant states

$$\dot{\psi} = 0, \dot{\theta} = 0 \tag{12}$$

$$\dot{\mathbf{l}}^b = 0 \tag{13}$$

$$\dot{f} = 0 \tag{14}$$

Equations (1)-(5) and (12)-(14) form the augmented state equations, with (10) being the measurement equation. Since the odometer provides velocity measurement in (10), the vehicle's position can only be acquired by integration of velocity and thus unobservable.

## III. Two Coupled Sub-problems and Observability Property

This section divides the problem of interest into two separate but tightly coupled sub-problems. Observability of the system is above all the first question to investigate, because if the system is unobservable, we have no way to achieve satisfactory estimation even if the measurement is perfect.

*Problem I: Self-calibration and Positioning*

It corresponds to the following simplified system

$$\dot{\mathbf{C}}_b^n = \mathbf{C}_b^n \left( \boldsymbol{\omega}_{nb}^b \times \right), \quad \boldsymbol{\omega}_{nb}^b = \boldsymbol{\omega}_{ib}^b - \mathbf{C}_n^b \boldsymbol{\omega}_{in}^n, \tag{15}$$

$$\dot{\mathbf{v}}^n = \mathbf{C}_b^n \mathbf{f}^b - \left( 2\boldsymbol{\omega}_{ie}^n + \boldsymbol{\omega}_{en}^n \right) \times \mathbf{v}^n + \mathbf{g}^n \tag{16}$$

$$\dot{\mathbf{p}} = \mathbf{R}_c \mathbf{v}^n \tag{17}$$

$$\dot{\psi} = 0, \dot{\theta} = 0 \tag{18}$$

$$\dot{\mathbf{l}}^b = 0 \tag{19}$$

$$\dot{f} = 0 \tag{20}$$

with the measurement equation given by (10). The sensor biases are left out because for a precision IMU they have negligible effect on the self-calibration result. For this system, we have the theorem that follows.

*Theorem 1 [10]*: If the vehicle runs on a trajectory with line and curving path segments, Problem I is partially observable. Specifically, the attitude and velocity is observable, the position is unobservable, the odometer lever arm $\mathbf{l}^b$ is observable, and the Euler angle and odometer factor triple $\begin{bmatrix} \psi & \theta & f \end{bmatrix}^T$ has four indiscriminable states, i.e.,

$$I\left\{ \begin{bmatrix} \psi & \theta & f \end{bmatrix}^T \right\} = \begin{Bmatrix} \begin{bmatrix} \pi+\psi & \pi-\theta & f \end{bmatrix}^T, \\ \begin{bmatrix} \pi+\psi & -\theta & -f \end{bmatrix}^T, \\ \begin{bmatrix} \psi & \pi+\theta & -f \end{bmatrix}^T \end{Bmatrix}, \quad -\pi < \psi, \theta \le \pi \tag{21}$$

The readers are referred to [10] for a rigorous proof. Therein the case with the absence of the odometer was also considered. The expression $I\{x_1\} = \{x_2\}$ means that a state $x_1$ is indiscriminable from another state $x_2$ in that they yield the same output. The indiscriminable (finite and discrete) states in (21) arise because of four different physical definitions of the vehicle frame and are thus all correct resolutions. Mathematically, this phenomenon appears as a result of the two unsigned 'zeros' in the nonholonomic constraint (9). It is interesting that the initial state of an estimator will automatically decide which definition to use among the four [11].

*Problem II: In-motion Alignment and Positioning*

On the condition of parameters in (12)-(14) being known, Problem II corresponds to another simplified system with the system equation given by (1)-(5), and the measurement equation by

$$\mathbf{y} = \mathbf{C}_n^b \mathbf{v}^n + \boldsymbol{\omega}_{eb}^b \times \mathbf{l}^b \tag{22}$$

where $\mathbf{y} \triangleq \frac{y_{odo}}{f}\left[\cos\theta\cos\psi \quad \sin\theta \quad -\cos\theta\sin\psi\right]^T$ is the body-referenced velocity of the center of the rear axle, which is available using the calibrated parameters.

*Theorem 2*: Problem II is observable in attitude and velocity, but unobservable in position.

Proof: Rewrite (22) as

$$\mathbf{C}_b^n\left(\mathbf{y} - \boldsymbol{\omega}_{eb}^b \times \mathbf{l}^b\right) = \mathbf{v}^n \tag{23}$$

Substituting into (2) and reorganizing the terms

$$\left(\boldsymbol{\omega}_{ib}^b - \mathbf{b}_g + \boldsymbol{\omega}_{ie}^b\right)\times\left(\mathbf{y} - \boldsymbol{\omega}_{eb}^b \times \mathbf{l}^b\right) + \dot{\mathbf{y}} - \dot{\boldsymbol{\omega}}_{eb}^b \times \mathbf{l}^b - \mathbf{f}^b + \mathbf{b}_a = \mathbf{C}_n^b \mathbf{g}^n \tag{24}$$

According to the chain rule of the attitude matrix, $\mathbf{C}_b^n$ at any time satisfies

$$\mathbf{C}_b^n(t) = \mathbf{C}_{b(t)}^{n(t)} = \mathbf{C}_{n(0)}^{n(t)}\mathbf{C}_{b(0)}^{n(0)}\mathbf{C}_{b(t)}^{b(0)} = \mathbf{C}_{n(0)}^{n(t)}\mathbf{C}_b^n(0)\mathbf{C}_{b(t)}^{b(0)} \tag{25}$$

where the initial attitude matrix $\mathbf{C}_b^n(0)$ is constant, and $\mathbf{C}_{b(0)}^{b(t)}$ and $\mathbf{C}_{n(0)}^{n(t)}$, respectively, encode the attitude changes of the body frame and the navigation frame from time $0$ to $t$. The attitude matrix $\mathbf{C}_{b(0)}^{b(t)}$ can be obtained by integrating the gyroscope output $\boldsymbol{\omega}_{ib}^b$, where the adverse effect of the gyroscope bias is negligible for a high-accuracy IMU, while the attitude matrix $\mathbf{C}_{n(0)}^{n(t)}$ can be computed by considering only the Earth rotation in this case of a land vehicle.

Substituting (25) into (24) gives

$$\begin{aligned}&\mathbf{C}_{b(t)}^{b(0)}\left[\left(\boldsymbol{\omega}_{ib}^b - \mathbf{b}_g + \boldsymbol{\omega}_{ie}^b\right)\times\left(\mathbf{y} - \boldsymbol{\omega}_{eb}^b \times \mathbf{l}^b\right) + \dot{\mathbf{y}} - \dot{\boldsymbol{\omega}}_{eb}^b \times \mathbf{l}^b - \mathbf{f}^b + \mathbf{b}_a\right] \\ &= \mathbf{C}_n^b(0)\mathbf{C}_{n(t)}^{n(0)}\mathbf{g}^n\end{aligned} \tag{26}$$

The Earth rotation rate $\boldsymbol{\omega}_{ie}^b$ and the gyroscope bias $\mathbf{b}_g$ are a small quantities as compared to the body angular rate $\boldsymbol{\omega}_{ib}^b$, so is the accelerometer bias $\mathbf{b}_a$ with respect to the specific force $\mathbf{f}^b$. The above equation is approximated as

$$\begin{aligned}&\mathbf{C}_{b(t)}^{b(0)}\left[\boldsymbol{\omega}_{ib}^b \times\left(\mathbf{y} - \boldsymbol{\omega}_{ib}^b \times \mathbf{l}^b\right) + \dot{\mathbf{y}} - \dot{\boldsymbol{\omega}}_{ib}^b \times \mathbf{l}^b - \mathbf{f}^b\right] \\ &= \mathbf{C}_n^b(0)\mathbf{C}_{n(t)}^{n(0)}\mathbf{g}^n\end{aligned} \tag{27}$$

The attitude matrix can be determined if the coordinates of any two linearly independent vectors are given. In this case, $\mathbf{C}_{n(t)}^{n(0)}\mathbf{g}^n$ is the gravity vector seen from the inertial frame and its time history normally forms a cone (cf. Fig. 1 in [15]), so there always exist two time instants so that $\mathbf{C}_{n(t)}^{n(0)}\mathbf{g}^n$ have linearly independent directions. Now the initial attitude matrix $\mathbf{C}_b^n(0)$ is known, so is the attitude matrix at any time $\mathbf{C}_b^n$ by (25). Consequently, the velocity is known according to (23).

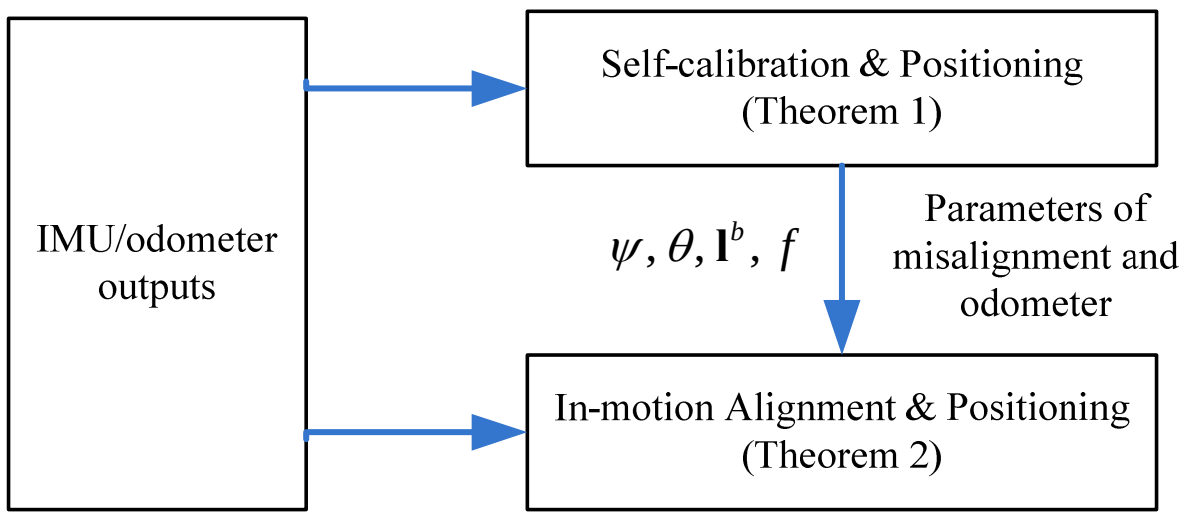

Figure 2. Feasibility of self-calibration, in-motion alignment and positioning for IMU/odometer system.

Theorems 1-2 collectively state that, under very moderate conditions, it is feasible to self-calibrate parameters of the IMU/odometer system and fulfill the advantageous functionality of in-motion alignment and positioning. Of course, position will suffer from an accumulating error drift since it is unobservable. Special technique should be taken to mitigate the position drift. This relationship is depicted in Fig. 2.

## IV. Algorithms and Implementation

A unified solution to the problems of self-calibration, in-motion alignment and accurate positioning is extremely difficult, if not impossible. The paper tends to divide and conquer them making use of the special characteristics of the land vehicle and the precision IMU. Specifically, as for the sub-problem of self-calibration and positioning, it is quite convenient to let the vehicle start to go from zero velocity and the self-calibration task will benefit from the ground static alignment that can provide a good attitude initialization. For the second sub-problem, we assume the parameters in (12) -(14) have been known and fixed during the in-motion alignment, and then refine them for better navigation performance immediately after the in-motion alignment. For both sub-problems, we resort to the indirect extended Kalman filter (EKF) as the main workhorse to carry out the tasks [2]. Figure 3 gives the implementation schemes for the two coupled problems.

*Problem I: Self-calibration and Positioning*

We let the land vehicle start to run from zero velocity at a known position. This case allows a good attitude initialization and the EKF implementation is straightforward. The filter state comprises the vehicle's attitude, velocity and position; the gyroscope/accelerometer biases; and the misalignment and odometer parameters to be calibrated.

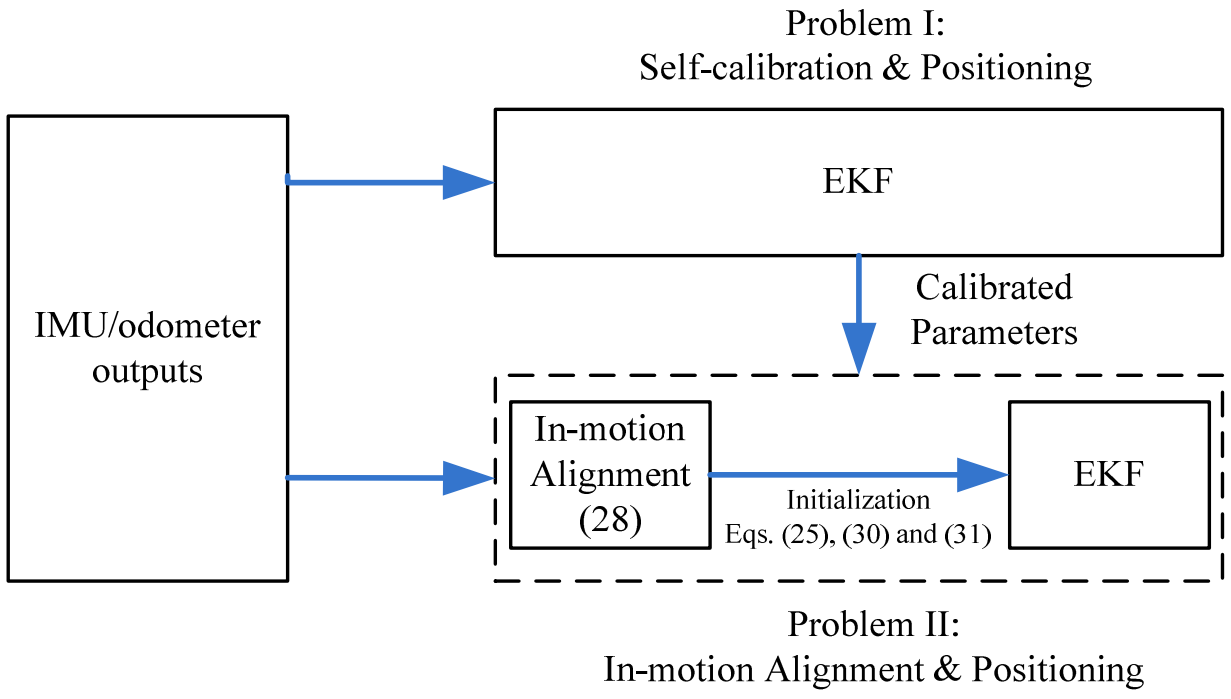

Figure 3. Implementation schemes of two coupled problems.

Table I. In-motion Alignment: odometer-aided vs. GPS-aided

| | Velocity Reference | Basic Equation |
|---|---|---|
| **Odometer-aided** | (local) body frame | (27): $\mathbf{C}_{b(t)}^{b(0)}\left[\boldsymbol{\omega}_{ib}^{b}\times\left(\mathbf{y}-\boldsymbol{\omega}_{ib}^{b}\times\mathbf{l}^{b}\right)+\dot{\mathbf{y}}-\dot{\boldsymbol{\omega}}_{ib}^{b}\times\mathbf{l}^{b}-\mathbf{f}^{b}\right]=\mathbf{C}_{n}^{b}\left(0\right)\mathbf{C}_{n(t)}^{n(0)}\mathbf{g}^{n}$ |
| **GPS-aided** | (global) Earth frame | (8) in [1]: $\mathbf{C}_{b(t)}^{b(0)}\mathbf{f}^{b}=\mathbf{C}_{n}^{b}\left(0\right)\mathbf{C}_{n(t)}^{n(0)}\left[\dot{\mathbf{v}}^{n}+\left(2\boldsymbol{\omega}_{ie}^{n}+\boldsymbol{\omega}_{en}^{n}\right)\times\mathbf{v}^{n}-\mathbf{g}^{n}\right]$ |

*Problem II: In-motion Alignment and Positioning*

For this case, the on-the-run IMU attitude initialization is the most critical stage to make sure a normal system starting-up, as well as accurate positioning. Next we present an odometer-aided IMU alignment method that is similar in spirit to our previous approach in the context of airborne alignment in [1]. The odometer-aided alignment is relatively slow because the velocity measurement is provided in the local INS body frame, in contrast to the global Earth frame in the GPS-aided alignment [1]. The frame where the velocity is referenced makes a big difference in terms of alignment speed. Specifically, Table I compares the odometer-aided alignment with the GPS-aided alignment. The former largely depends on Earth rotation to form independent directions ($\mathbf{C}_{n(t)}^{n(0)}\mathbf{g}^{n}$), while the latter could resort to additional motion maneuvers (the velocity-related terms in $\mathbf{C}_{n(t)}^{n(0)}\left[\dot{\mathbf{v}}^{n}+\left(2\boldsymbol{\omega}_{ie}^{n}+\boldsymbol{\omega}_{en}^{n}\right)\times\mathbf{v}^{n}-\mathbf{g}^{n}\right]$) to remarkably speed up this process.

Assume the vehicle starts the in-motion alignment at a known position. Integrating (27) over the time interval of interest $\left[0,\ t\right]$, we have

$$\begin{aligned}&\mathbf{C}_{b(t)}^{b(0)}\left(\mathbf{y}-\boldsymbol{\omega}_{ib}^{b}\times\mathbf{l}^{b}\right)-\left(\mathbf{y}\left(0\right)-\boldsymbol{\omega}_{ib}^{b}\left(0\right)\times\mathbf{l}^{b}\right)-\int_{0}^{t}\mathbf{C}_{b(t)}^{b(0)}\mathbf{f}^{b}dt\\&=\mathbf{C}_{n}^{b}\left(0\right)\int_{0}^{t}\mathbf{C}_{n(t)}^{n(0)}\mathbf{g}^{n}dt\end{aligned}\qquad(28)$$

Suppose the current time $t$ is an integer, say $M$, times of the updated interval, i.e., $t\triangleq MT$, where $T$ is the uniform time duration of the update interval $\left[t_{k}\quad t_{k+1}\right]$ with $t_{k}=kT$. The above two integrals can be calculated using the technique in [1] as

$$\begin{aligned}&\int_{0}^{t}\mathbf{C}_{n(t)}^{n(0)}\mathbf{g}^{n}dt\approx\sum_{k=0}^{M-1}\mathbf{C}_{n(t_{k})}^{n(0)}\left(T\mathbf{I}+\frac{T^{2}}{2}\boldsymbol{\omega}_{in}^{n}\times\right)\mathbf{g}^{n}\\&\int_{0}^{t}\mathbf{C}_{b(t)}^{b(0)}\mathbf{f}^{b}dt\approx\sum_{k=0}^{M-1}\mathbf{C}_{b(t_{k})}^{b(0)}\left[\begin{aligned}&\Delta\mathbf{v}_{1}+\Delta\mathbf{v}_{2}+\frac{1}{2}\left(\Delta\boldsymbol{\theta}_{1}+\Delta\boldsymbol{\theta}_{2}\right)\times\left(\Delta\mathbf{v}_{1}+\Delta\mathbf{v}_{2}\right)\\&+\frac{2}{3}\left(\Delta\boldsymbol{\theta}_{1}\times\Delta\mathbf{v}_{2}+\Delta\mathbf{v}_{1}\times\Delta\boldsymbol{\theta}_{2}\right)\end{aligned}\right]\end{aligned}\qquad(29)$$

where $\Delta\mathbf{v}_{1},\Delta\mathbf{v}_{2}$ are the first and second samples of the accelerometer-measured incremental velocity and $\Delta\boldsymbol{\theta}_{1},\Delta\boldsymbol{\theta}_{2}$ are the first and second samples of the gyroscope-measured incremental angle, respectively, during the update interval $\left[t_{k}\quad t_{k+1}\right]$. The initial attitude matrix $\mathbf{C}_{b}^{n}\left(0\right)$ can be obtained from (28) by solving a problem of minimum eigenvector as done in [1].

The alignment result, namely, the attitude matrix at any time $\mathbf{C}_{b}^{n}$ by (25), could then be used to set the initial attitude of an EKF for Problem II. The EKF here is roughly the same as that for Problem I, except the initial state and covariance settings. The EKF's initial velocity is calculated using (23), i.e.,

$$\mathbf{v}^{n}\approx\mathbf{C}_{b}^{n}\left(\mathbf{y}-\boldsymbol{\omega}_{ib}^{b}\times\mathbf{l}^{b}\right)\qquad(30)$$

At the very end of in-motion alignment, the vehicle usually displaces far from the known starting position. To restore the vehicle's position at current time, we need to save the history data of odometer outputs $y_{odo}$ and two inertially referenced attitude matrix, namely, $\mathbf{C}_{b(0)}^{b(t)}$ and $\mathbf{C}_{n(0)}^{n(t)}$. Then once the alignment period terminates, the vehicle's position at current time can be approximately restored as such

$$\mathbf{p}\approx\mathbf{p}\left(0\right)+\mathbf{C}_{b}^{n}\left(0\right)\mathbf{l}^{b}+T\sum_{k=0}^{M-1}\mathbf{C}_{n(0)}^{n(t_{k})}\mathbf{C}_{b}^{n}\left(0\right)\mathbf{C}_{b(t_{k})}^{b(0)}\mathbf{y}-\mathbf{C}_{b}^{n}\mathbf{l}^{b}\qquad(31)$$

This technique was also used in [16] that disregarded the translational misalignment.

## V. Test Results

The land vehicle is equipped with a navigation-grade ringer laser gyroscope (RLG) IMU and an odometer installed between the gear box and the back axle. The IMU body frame is roughly aligned with backward, upward, left directions of the vehicle. The test data was collected on a round trip from the Changsha city to the Changde city, China. The distance between the two cities is about 200 km. The outbound-trip (Changsha-Changde) and inbound-trip (Changde-Changsha) datasets were collected separately since they took place with an idle day in between. GPS data was also collected as a comparison reference. The odometer measures in pulses the incremental distance along the track. We use a linear Kalman pre-filter running at the same rate as the raw data to calculate the instantaneous speed from the incremental pulses, assuming an along-track const-acceleration dynamic model. The filtered speed is fed to the EKFs as measurements at 1Hz.

The self-calibration test starts to work following a stationary alignment lasting 300 seconds. Figures 4-6 present the calibration results for both the outbound-trip dataset and the inbound-trip dataset, Fig. 7 plots the estimation results of the gyroscope and accelerometer biases and Figs. 8-9 plot the positioning errors relative to the distance traveled. The trajectory reference is generated using INS/GPS integration

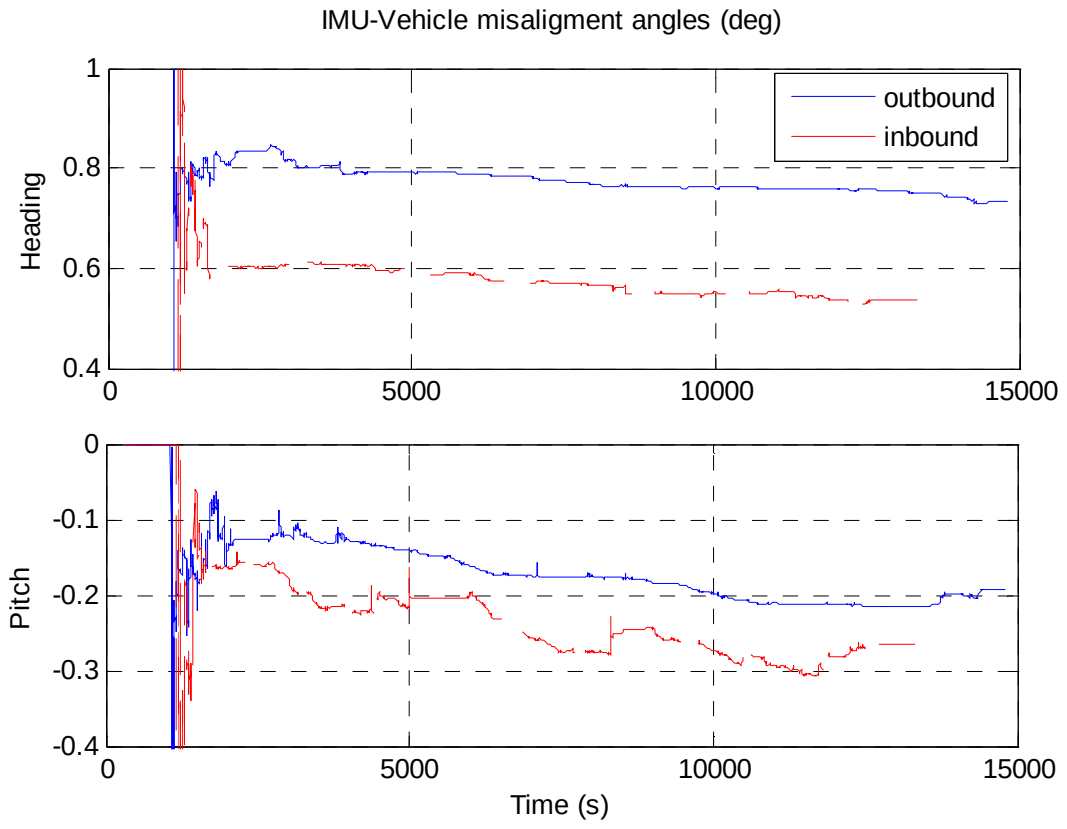

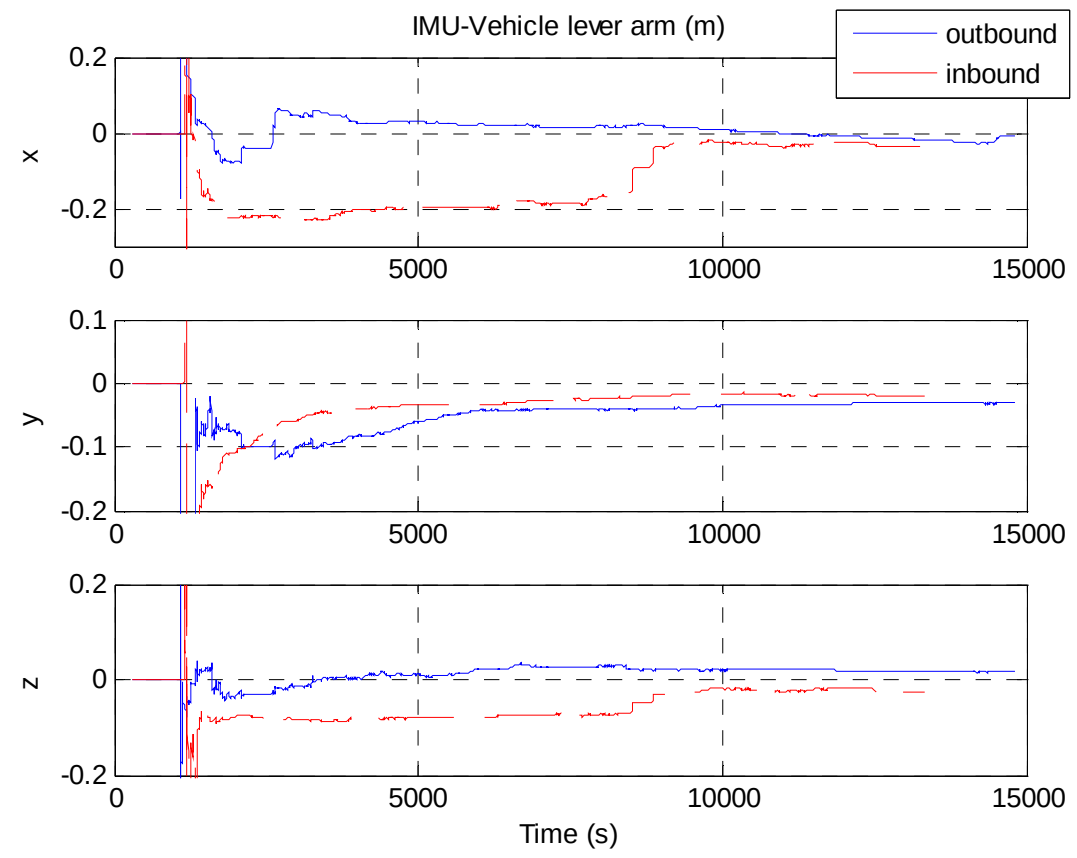

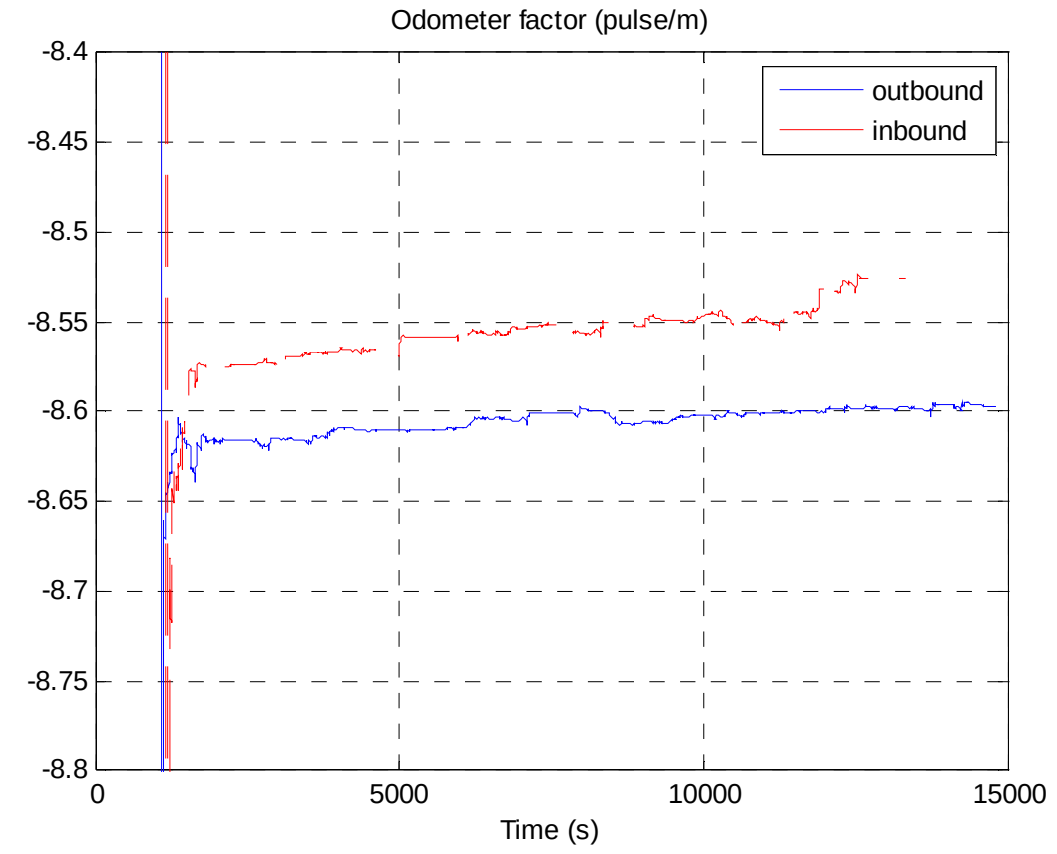

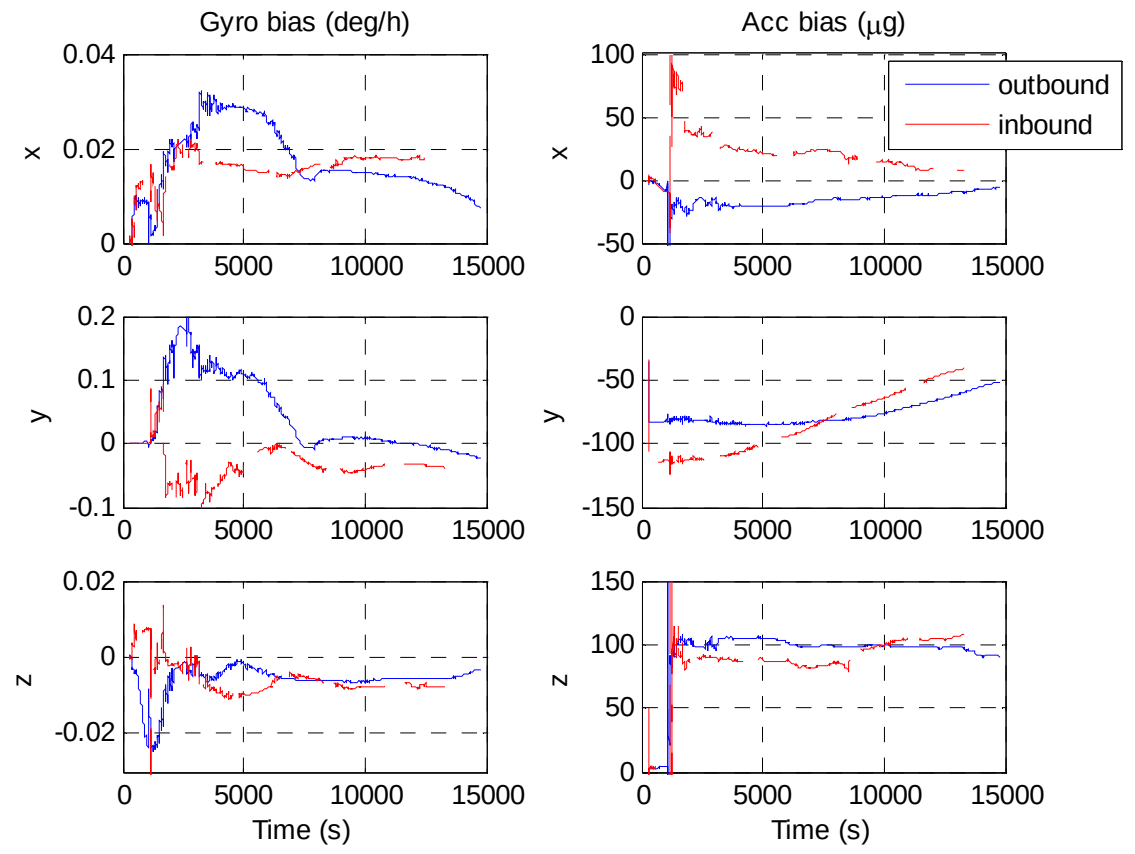

Figure 7. Estimation results of gyroscope and accelerometer biases on a two-way trip.

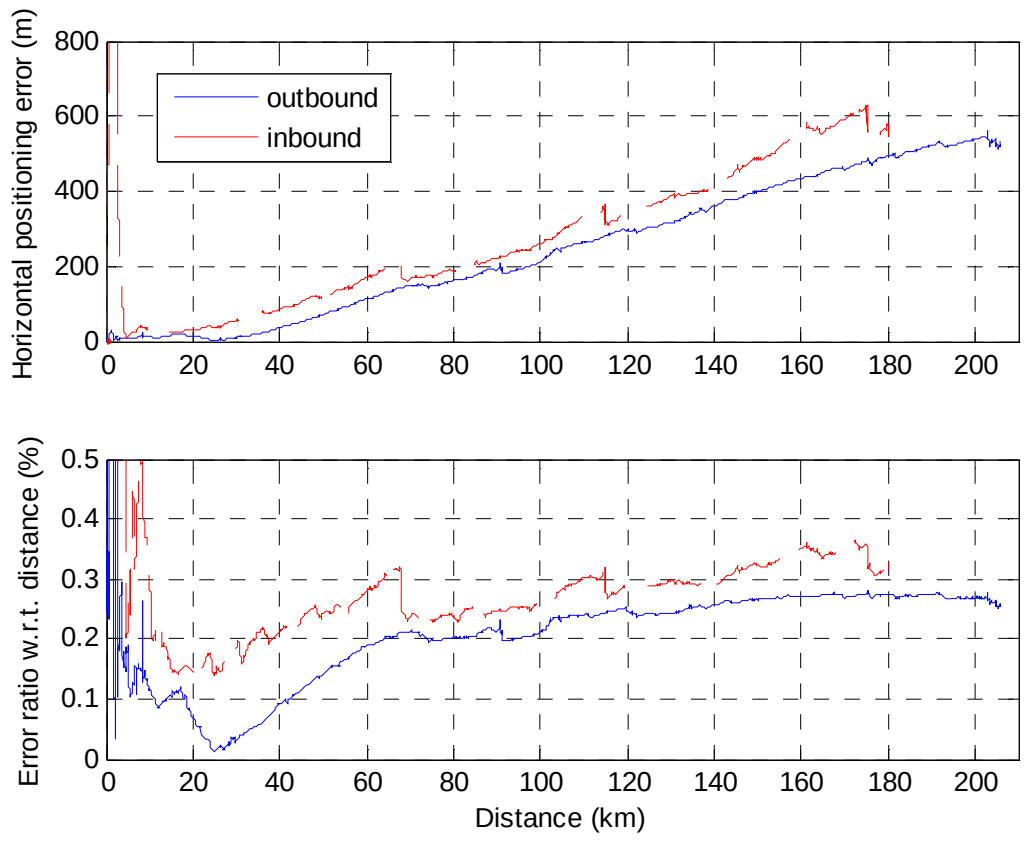

Figure 8. Horizontal positioning errors relative to distance travelled on a two-way trip.

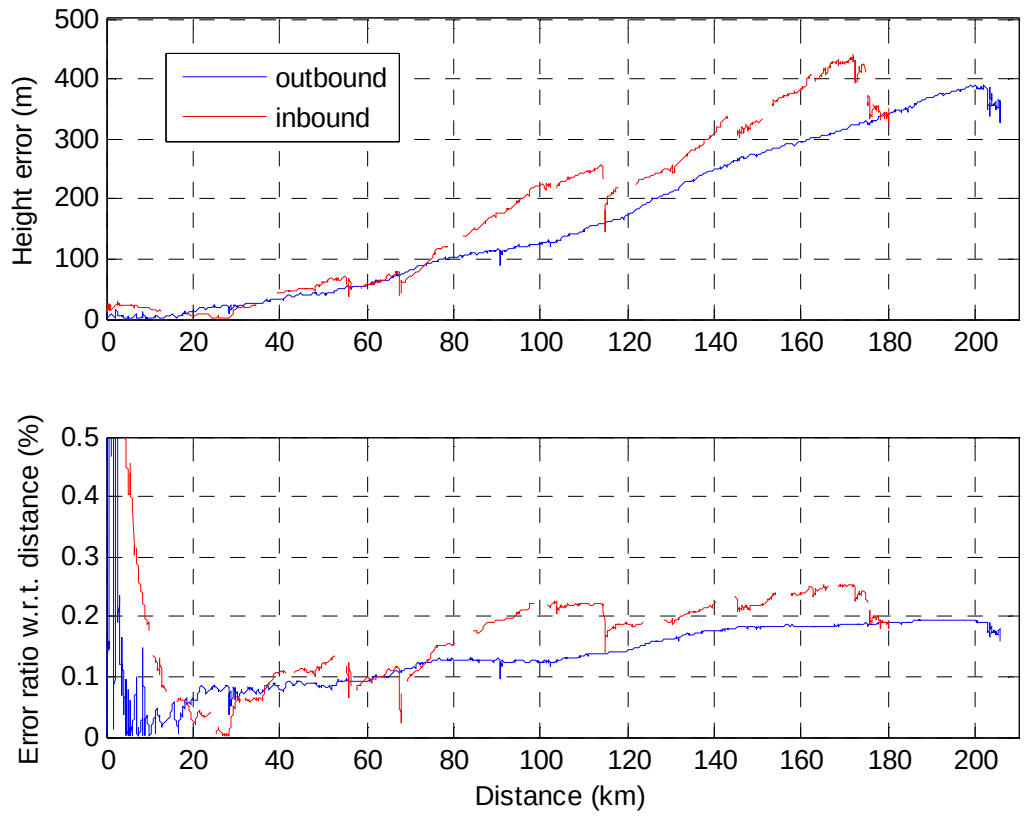

Figure 9. Height positioning errors relative to distance travelled on a two-way trip

[12]. The back trip begins with a gravel road, which adversely affects its positioning accuracy. It contributes to hundreds of horizontal positioning errors over the first several kilometers in the inbound trip (see Fig. 8). Taking this affect into account, it is reasonable to claim that the horizontal positioning accuracy is better than 0.3%D and the height accuracy is better than 0.2%D, where D means the distance travelled. The odometer results in Fig. 6 show an obvious deceasing trend in magnitude on both trips. It is because the tires inflate along with the gradually increasing tire temperature due to land friction, resulting in less pulses for each meter of travel. In addition, the odometer scale factor result on the inbound trip (~8.55 pulses/meter) is 0.6% smaller in magnitude than that on the outbound trip (~8.6 pulses/meter). This significant variation amount is caused by different air temperature when the two trips took place, namely, 18℃ (outbound trip) versus 26℃ (inbound trip). It is likely that the tire inflation variation leads to the discrepancy of IMU-vehicle misalignment angles, as seen in

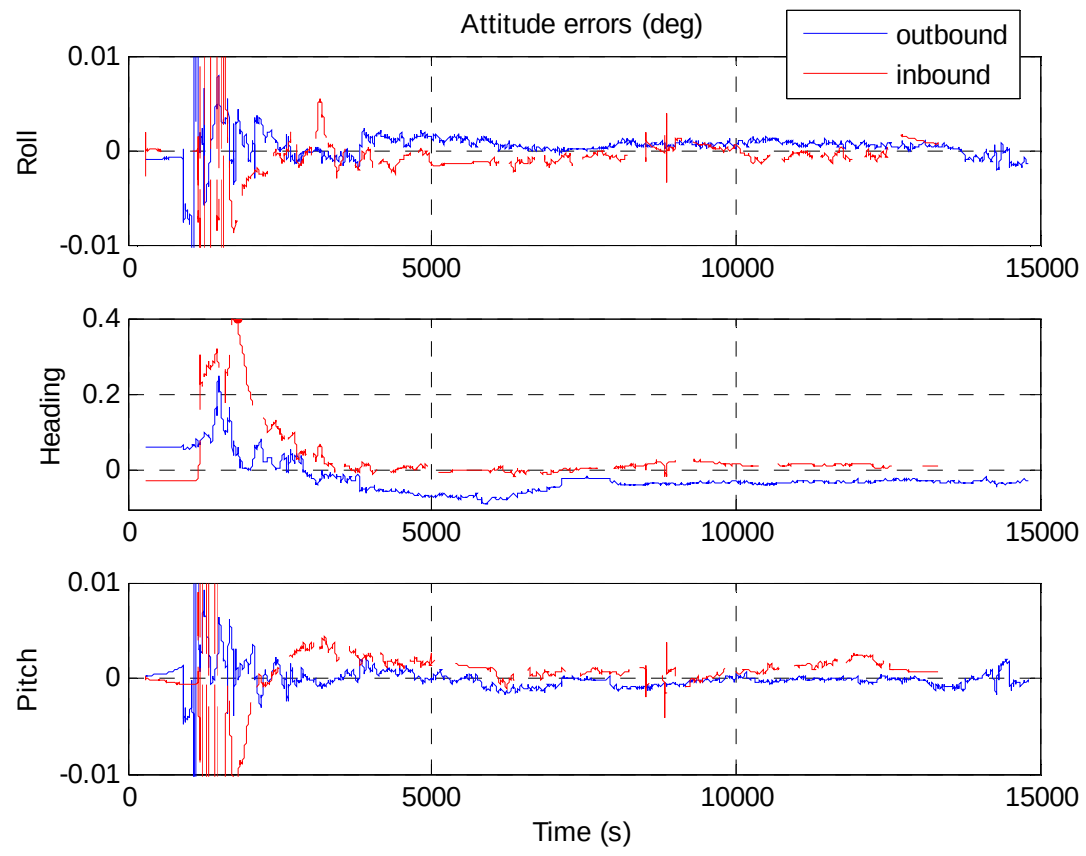

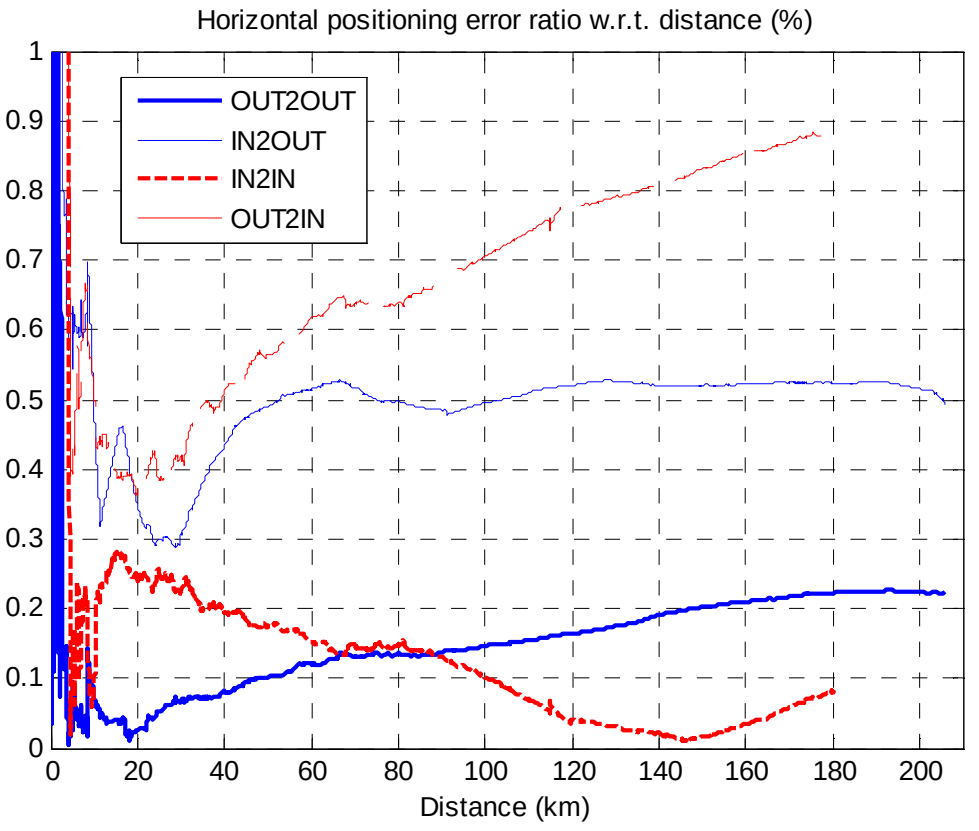

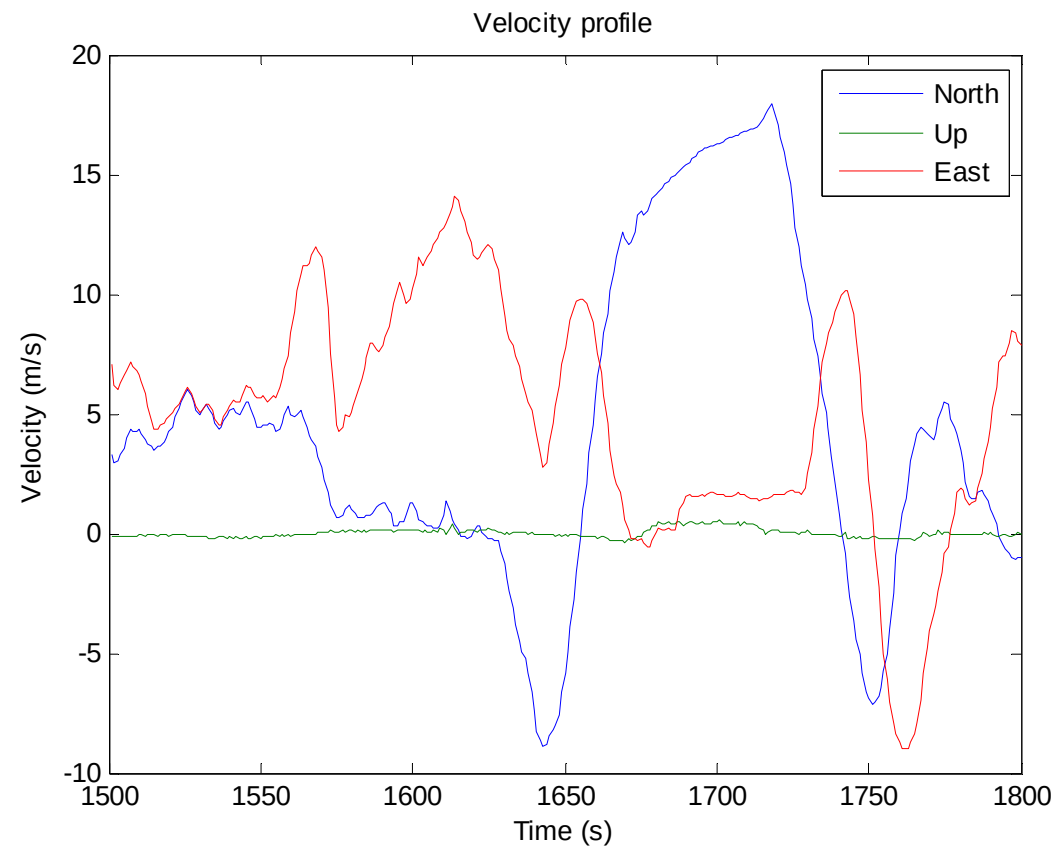

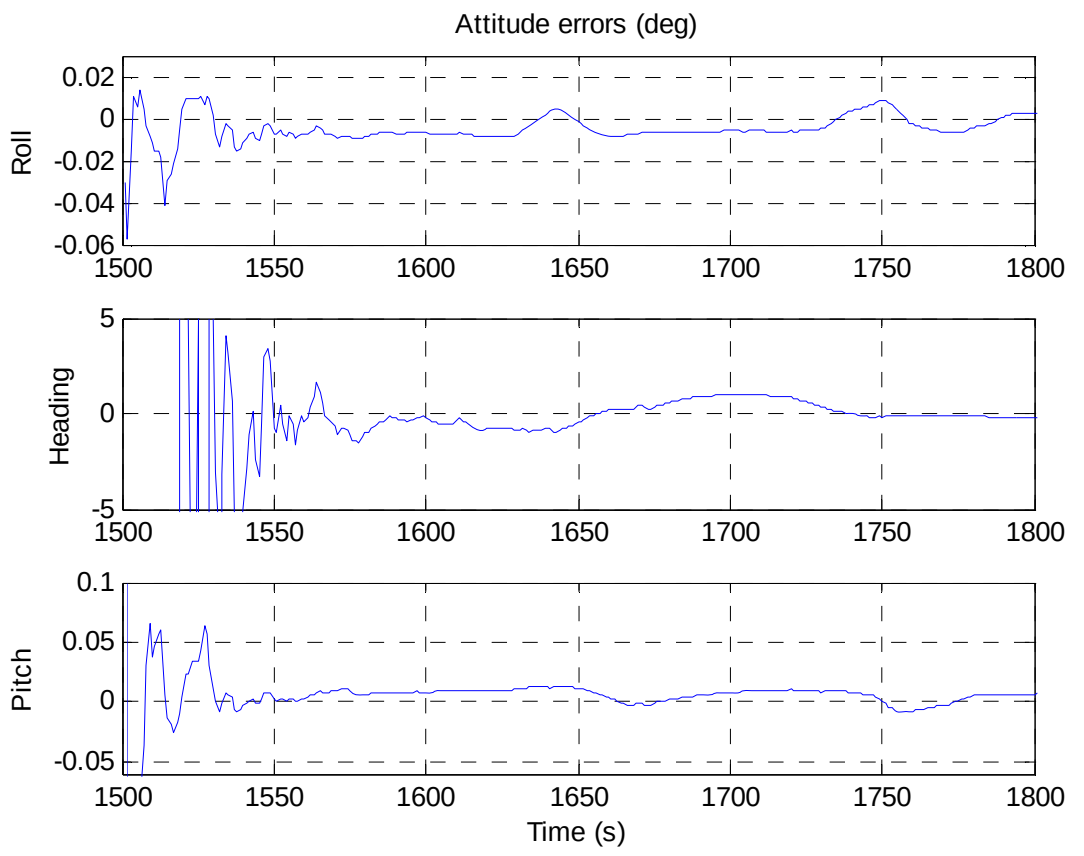

Figure 13. Attitude errors during in-motion alignment.

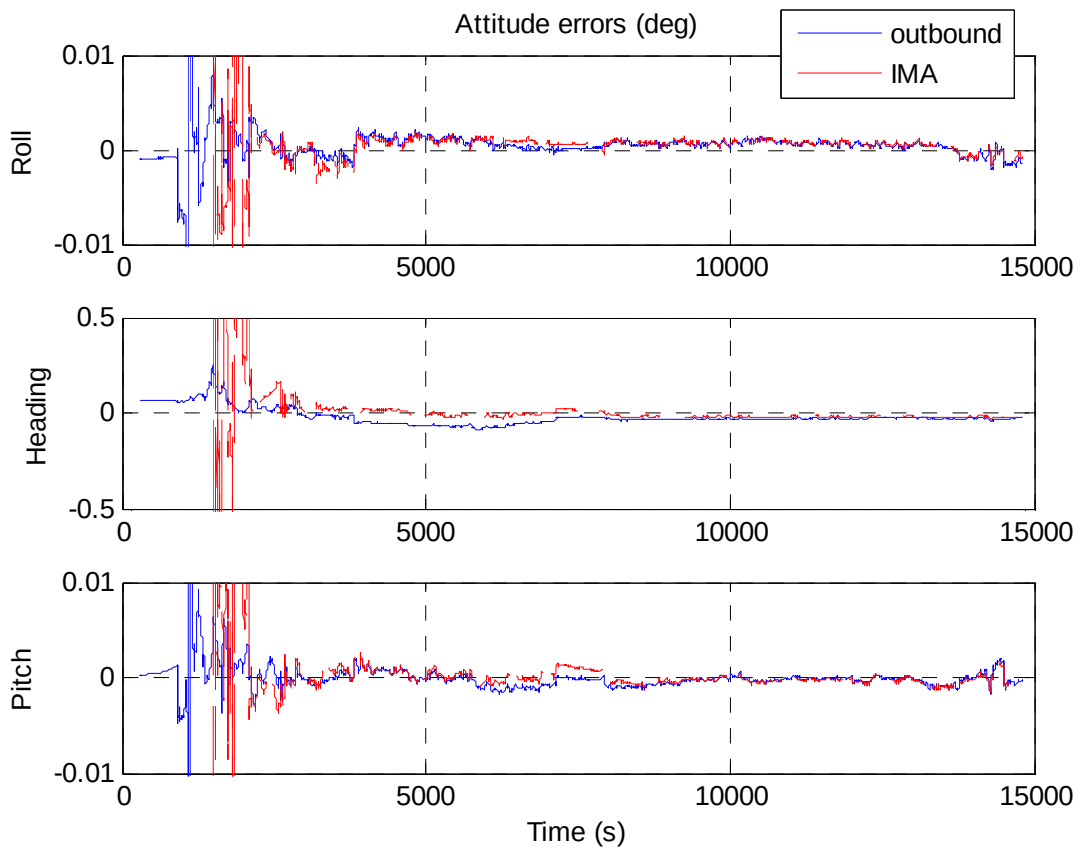

Figure 14. Attitude errors on the outbound trip with in-motion alignment (indicated by IMA), as compared with those by self-calibration.

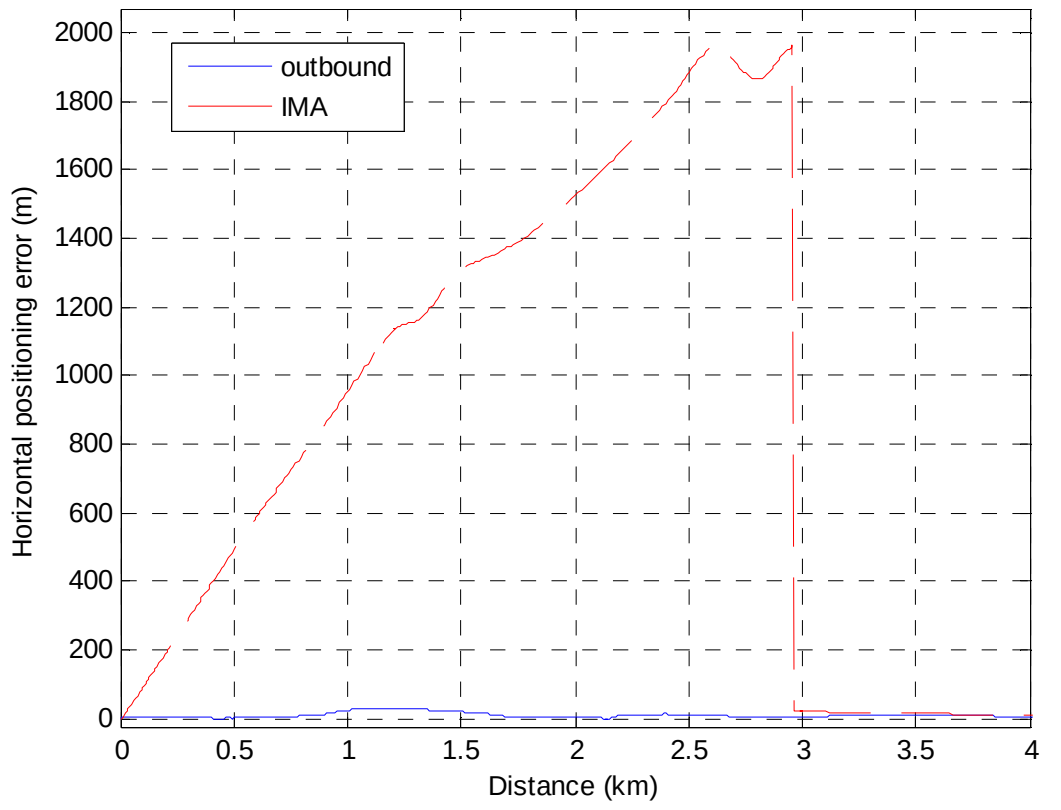

Figure 15. Horizontal positioning error during and right after in-motion alignment (indicated by IMA), as compared with that by self-calibration.

Fig. 4. On the other hand, the proposed self-calibration and positioning algorithm is proved having successfully mitigated as large variation of the scale factor as 0.6%, to reach a position accuracy of 0.2-0.3%D. By inspecting the attitude errors relative to the INS/GPS reference (see Fig. 10), we see that although no global measurement is provided at all, a high attitude accuracy is maintained by integrating IMU and the odometer throughout the trips. This observation accords with our analysis in Theorem 1.

Now two set of calibrated parameters are respectively obtained from the separate outbound and inbound trips. Figures 4-6 show that they have non-negligible discrepancy. Although the true parameters are unknown, the correctness of the calibrated parameters can be verified by a cross-checking technique, that is to say, by fixing and applying each set of calibration parameters to the two test data. For this cross-checking verification, (10) still acts as the measurement equation, in which the parameters are fixed quantities as determined by the self-calibration process. Figure 10 presents the cross-checking result in terms of horizontal positioning accuracy. For example, 'OUT2IN' indicates the horizontal positioning result by applying the calibrated parameters on the outbound trip to the dataset of the inbound trip. We see that the two set of calibration parameters are good for their respective datasets, but poor when the parameters obtained from one dataset is applied to the other dataset. In view of the good accuracy for both trips in Fig. 8, we may conclude that environmental factors, such as temperature, have significant impact on the IMU-vehicle misalignment and odometer parameters, and the proposed self-calibration and positioning algorithm is capable to effectively depress these negative factors.

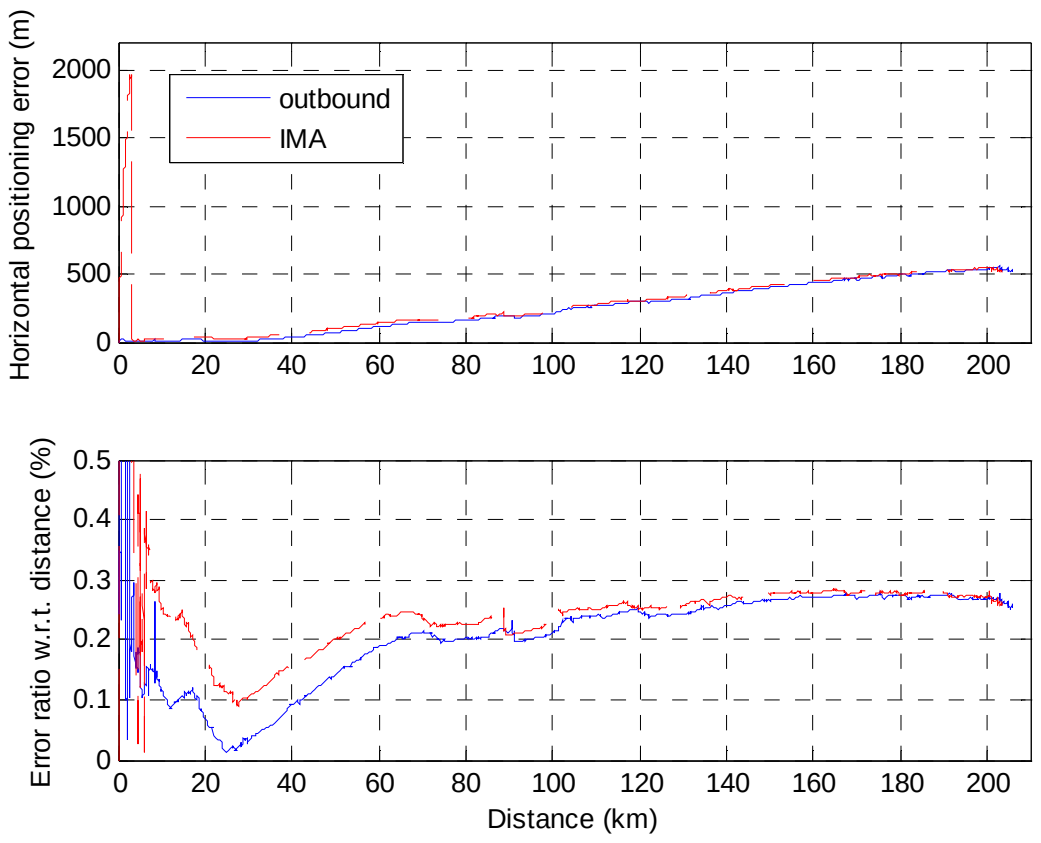

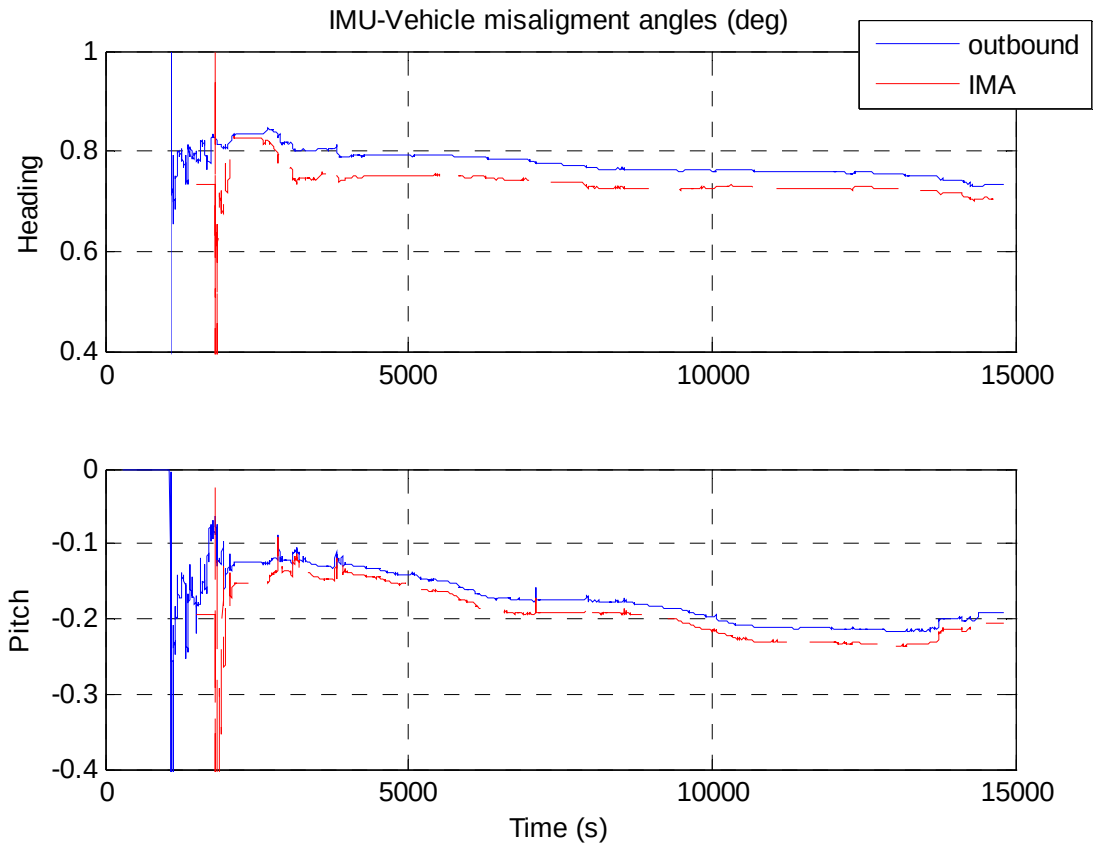

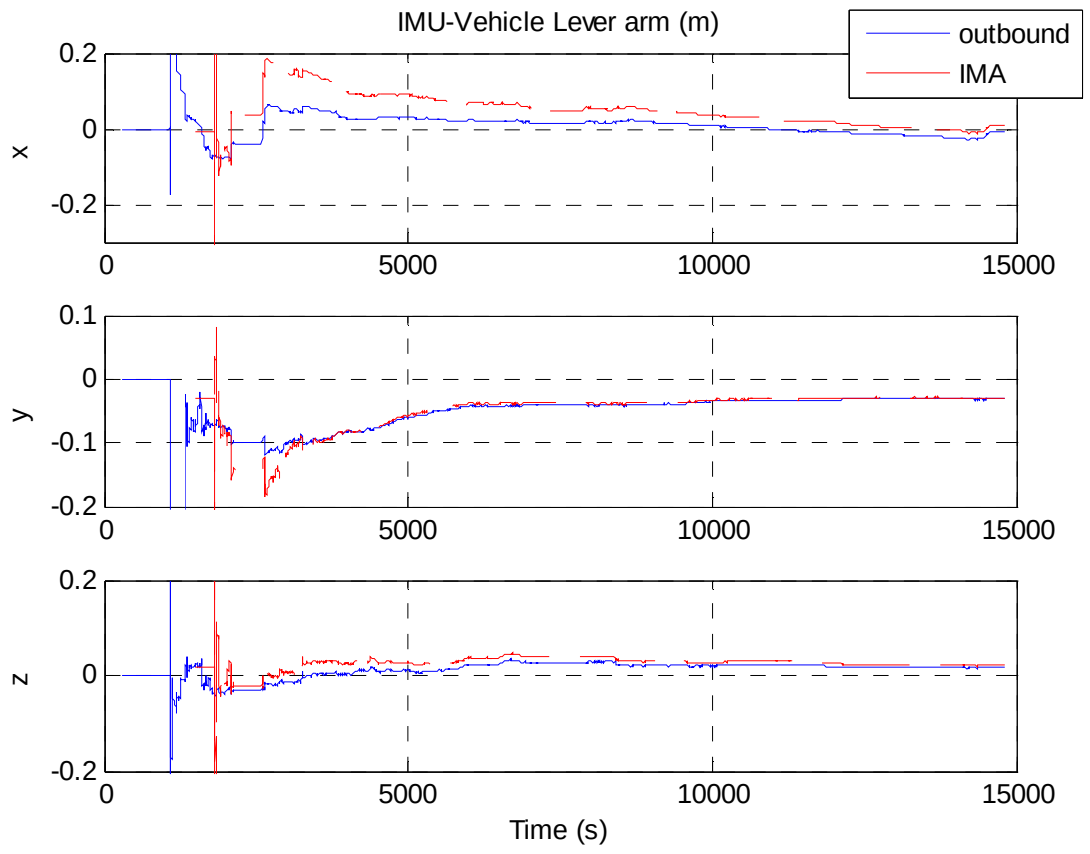

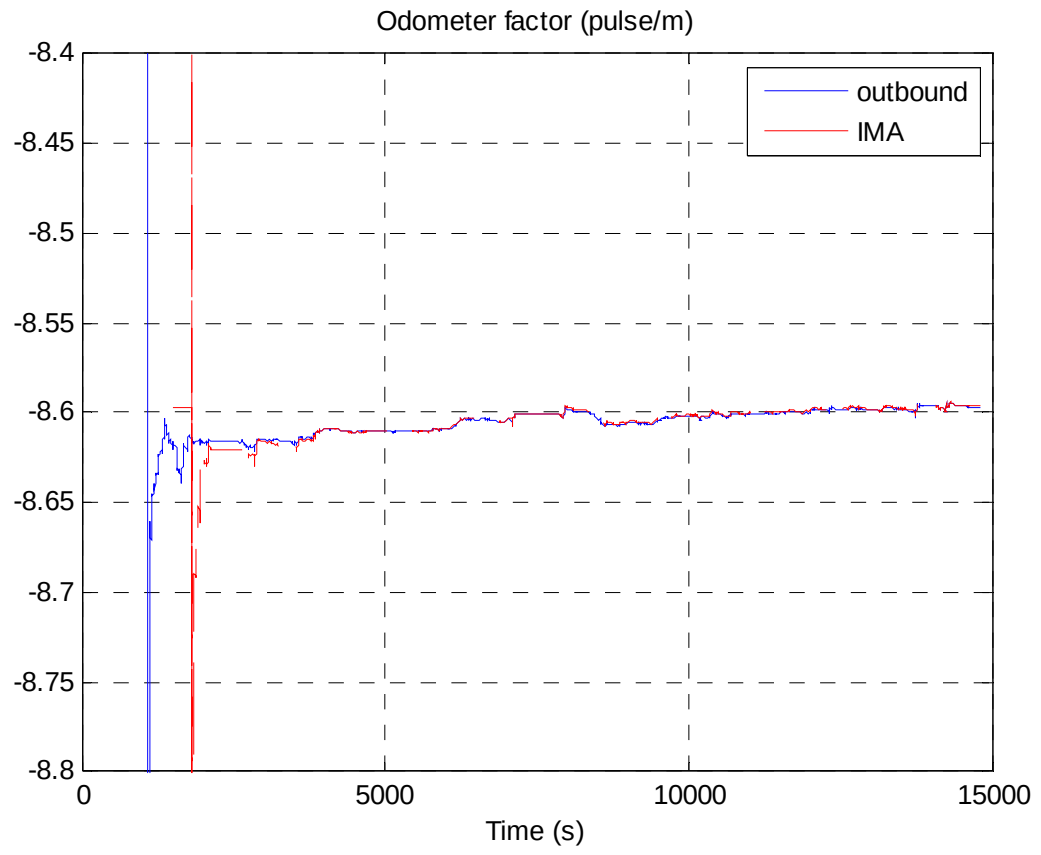

Figure 19. Estimate of odometer scale factor on the outbound trip with in-motion alignment (indicated by IMA), as compared with those by self-calibration.

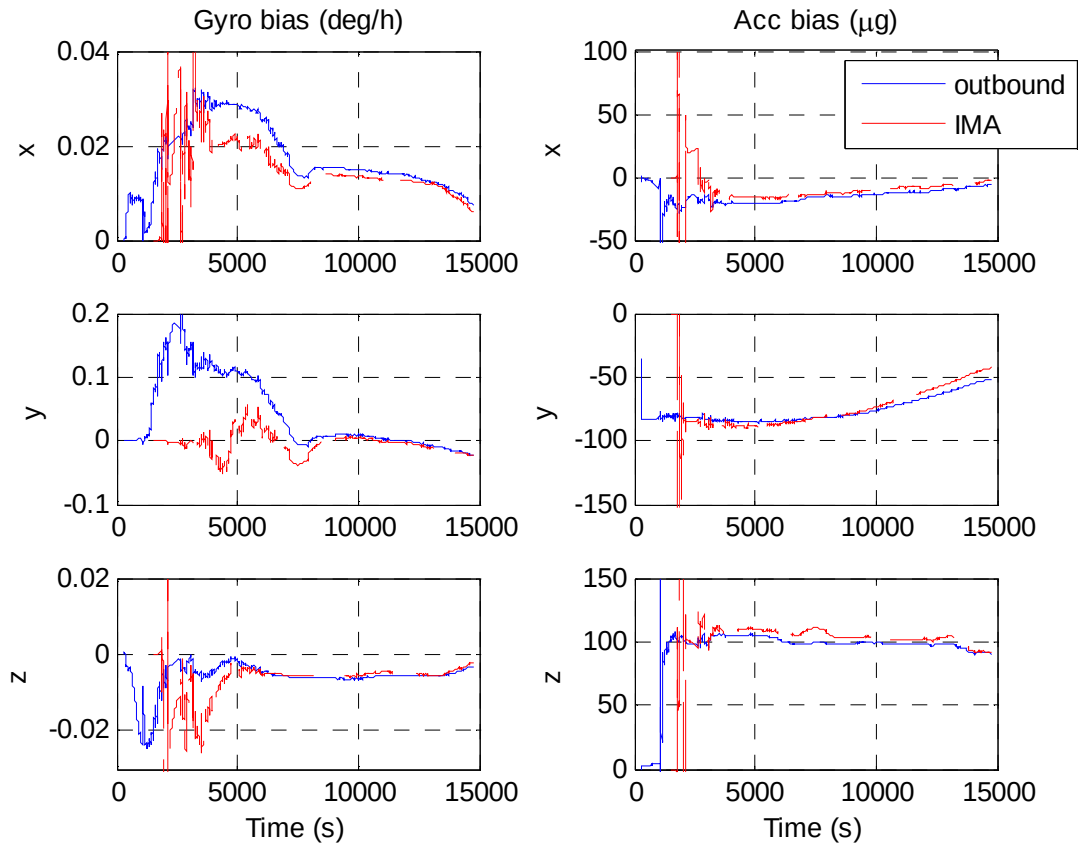

Figure 20. Estimate of gyroscope/accelerometer bias on the outbound trip with in-motion alignment (indicated by IMA), as compared with those by self-calibration.

For the case of Problem II, we present the test result below via the outbound-trip dataset. In-motion alignment is carried out for 300 seconds when the vehicle was running freely. The vehicle's velocity profile during the in-motion alignment is plotted in Fig.12 and the alignment angle errors are given in Fig. 13. It appears that the heading error reduces to within 1 deg in a couple of minutes, with the level angle errors within 0.01 deg. Note that this result has insignificant difference when the other set of calibrated parameters is used. Figure 14 gives the attitude errors over the whole outbound trip period and we see that the attitude accuracy with an in-motion alignment performs quite similarly with those by the self-calibration in Fig. 10. Figure 15 depicts the horizontal positioning error during and right after the in-motion alignment, in which the positioning error is effectively reduced by (31) from 2000 m to about 20 m. The positioning accuracy for the whole dataset (see Fig. 16) is better than 0.3%D, marginally worse than that by the self-calibration starting from still. The refined parameters, as well as the estimates of gyroscope/accelerometer biases, are presented in Figs. 17-20 and compared with those by the self-calibration. As expected, they accord with each other, which confirms the correctness of the implemented EKFs.

## VI. Discussions and Conclusions

A wheeled land vehicle cannot move freely because it is subjected to the nonholonomic constraint. This motion constraint, together with an odometer measuring the forward speed, could be effectively used to assist the INS onboard the vehicle, if only the spatial misalignment between the INS and the vehicle were correctly determined. Unfortunately, the misalignment problem has seldom been seriously addressed in the literature or by the famous navigation product providers. In fact, the misalignment problem is a major barrier to an accurate and autonomous land navigation system using inertial sensors and an odometer. This is not only because that the misalignment parameters may be difficult to accurately acquire, but also because they may vary significantly due to such adverse factors as ambient temperature and load changing, as reported in this paper. It is shown that these parameters cannot be regarded as constant, and instead should be calibrated or refined in each run so as to achieve a good navigation performance.

In this paper, we proposes a versatile strategy for self-contained land navigation using the IMU and an odometer. In view of the difficulty in solving the IMU/odometer integration as a whole, we design algorithms to solve two separate but tightly coupled sub-problems. Specifically, the algorithm of self-calibration and positioning applies when the vehicle starts to go from zero velocity. It is able to estimate the required parameters from scratch and meanwhile achieve good positioning. Given the calibrated parameters, the algorithm of in-motion alignment and positioning can work when the vehicle is running freely, and further refine those parameters to get good performance. This is made possible by the devised odometer-aided IMU alignment algorithm on the run. The results in the 200 km real tests prove the effectiveness of the proposed algorithms and show the potentials of the versatile land navigation strategy that might enable easy maintenance, real autonomy and high accuracy of the system.

## Acknowledgements

The authors appreciate the navigation group in Department of Automatic Control, National University of Defense Technology for providing the test data.